\documentclass[sigconf]{acmart}

\AtBeginDocument{%
  \providecommand\BibTeX{{%
    \normalfont B\kern-0.5em{\scshape i\kern-0.25em b}\kern-0.8em\TeX}}}

\setcopyright{acmcopyright}
\copyrightyear{2022}
\acmYear{2022}
\acmDOI{}

\newcount\COMMENTs  
\usepackage{color}
\definecolor{darkgreen}{rgb}{0,0.5,0}
\definecolor{purple}{rgb}{1,0,1}
\newcommand{\comm}[2]{\ifnum\COMMENTs=1\textcolor{#1}{#2}\fi}

\newcommand{\hide}[1]{}

\copyrightyear{2022}
\acmYear{2022}
\setcopyright{acmlicensed}
\acmConference[KDD '22] {Proceedings of the 28th ACM SIGKDD Conference on Knowledge Discovery and Data Mining}{August 14--18, 2022}{Washington, DC, USA.}
\acmBooktitle{Proceedings of the 28th ACM SIGKDD Conference on Knowledge Discovery and Data Mining (KDD '22), August 14--18, 2022, Washington, DC, USA}
\acmPrice{15.00}
\acmISBN{978-1-4503-9385-0/22/08}
\acmDOI{10.1145/3534678.3539045}

\newcommand{\xhdr}[1]{{\noindent\bfseries #1}.}

\def\G{\mathcal{G}}

\usepackage{makecell}
\usepackage{multicol}
\usepackage{subcaption}
\begin{document}

\title{Learning Large-scale Subsurface Simulations with a Hybrid Graph Network Simulator}

 \author{Tailin Wu}
 \email{tailin@cs.stanford.edu}
 \affiliation{
  \institution{Stanford University}
  \streetaddress{}
  \city{}
  \state{}
  \country{}
  \postcode{}
 }
 
 \author{Qinchen Wang}
 \email{qinchenw@cs.stanford.edu}
 \affiliation{
  \institution{Stanford University}
  \streetaddress{}
  \city{}
  \state{}
  \country{}
  \postcode{}
 }
 
 \author{Yinan Zhang}
 \email{yinanzy@cs.stanford.edu}
 \affiliation{
  \institution{Stanford University}
  \streetaddress{}
  \city{}
  \state{}
  \country{}
  \postcode{}
 }
 
 \author{Rex Ying}
 \email{rexying@cs.stanford.edu}
 \affiliation{
  \institution{Stanford University}
  \streetaddress{}
  \city{}
  \state{}
  \country{}
  \postcode{}
 }
 
 \author{Kaidi Cao}
 \email{kaidicao@cs.stanford.edu}
 \affiliation{
  \institution{Stanford University}
  \streetaddress{}
  \city{}
  \state{}
  \country{}
  \postcode{}
 }
 
  \author{Rok Sosi\v{c}}
 \email{rok@cs.stanford.edu}
 \affiliation{
  \institution{Stanford University}
  \streetaddress{}
  \city{}
  \state{}
  \country{}
  \postcode{}
 }
 
\author{Ridwan Jalali}
 \email{ridwan.jalali@aramco.com}
 \affiliation{
  \institution{Saudi Aramco}
  \streetaddress{}
  \city{}
  \state{}
  \country{}
  \postcode{}
 }
 
 \author{Hassan Hamam}
 \email{hassan.hamam@aramco.com}
 \affiliation{
  \institution{Saudi Aramco}
  \streetaddress{}
  \city{}
  \state{}
  \country{}
  \postcode{}
 }

 \author{Marko Maucec}
 \email{marko.maucec@aramco.com}
 \affiliation{
  \institution{Saudi Aramco}
  \streetaddress{}
  \city{}
  \state{}
  \country{}
  \postcode{}
 }
 
 \author{Jure Leskovec}
 \email{jure@cs.stanford.edu}
 \affiliation{
  \institution{Stanford University}
  \streetaddress{}
  \city{}
  \state{}
  \country{}
  \postcode{}
 }
 
\renewcommand{\shortauthors}{Tailin Wu, et al.}

\begin{abstract}
Subsurface simulations use computational models to predict the flow of fluids (e.g., oil, water, gas) through porous media. These simulations are pivotal in industrial applications such as petroleum production, where fast and accurate models are needed for high-stake decision making, for example, for well placement optimization and field development planning. Classical finite difference numerical simulators require massive computational resources to model large-scale real-world reservoirs. Alternatively, streamline simulators and data-driven surrogate models are computationally more efficient by relying on approximate physics models, however they are insufficient to model complex reservoir dynamics at scale.

Here we introduce Hybrid Graph Network Simulator (HGNS), which is a data-driven surrogate model for learning reservoir simulations of 3D subsurface fluid flows. To model complex reservoir dynamics at both local and global scale, HGNS consists of a subsurface graph neural network (SGNN) to model the evolution of fluid flows, and a 3D-U-Net to model the evolution of pressure. HGNS is able to scale to grids with millions of cells per time step, two orders of magnitude higher than previous surrogate models, and can accurately predict the fluid flow for tens of time steps (years into the future). Using an industry-standard subsurface flow dataset (SPE-10) with 1.1 million cells, we demonstrate that HGNS is able to reduce the inference time up to 18 times compared to standard subsurface simulators, and that it outperforms other learning-based models by reducing long-term prediction errors by up to 21\%. Project website can be found at \url{http://snap.stanford.edu/hgns/}.
\end{abstract}

\keywords{subsurface simulations, hybrid graph neural network, multi-scale, large-scale}

\maketitle

\begin{figure}[!ht]
    \centering
      \includegraphics[width=1\linewidth]{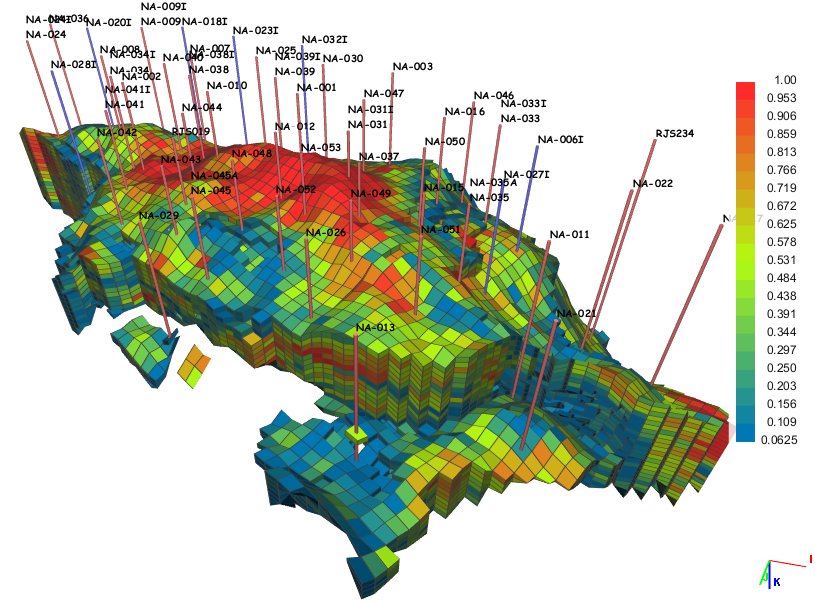}
    \caption{Subsurface simulation model \cite{abraham2009distributed}. Water and oil exist in the porous rock which is discretized into cells, representing a computational grid. To produce oil under waterflooding scheme, water is pumped into the reservoir using injectors (blue pins), which creates a pressure gradient and preferential fluid flow path from the injectors to the oil producers (red pins). Colors indicate oil saturation level.  }
    \label{fig:subsurface}
\end{figure}

\section{Introduction}
Subsurface simulation is a discipline, where computational models are used to predict the flow of fluids (e.g., oil, water, gas) through porous media. It is pivotal for effective management of hydrocarbon and groundwater resources. In the petroleum industry, reservoir simulation is essential for efficient
oil and gas field development planning, where decisions on well placement and well management require fast and reliable subsurface
simulation models that are dynamically calibrated, i.e., history matched. Such reservoir models reproduce historical operational events and facilitate production forecasting and optimization under reservoir uncertainty \cite{dogru2011new}.

\begin{figure*}[!ht]
    \centering
      \includegraphics[width=0.75\linewidth]{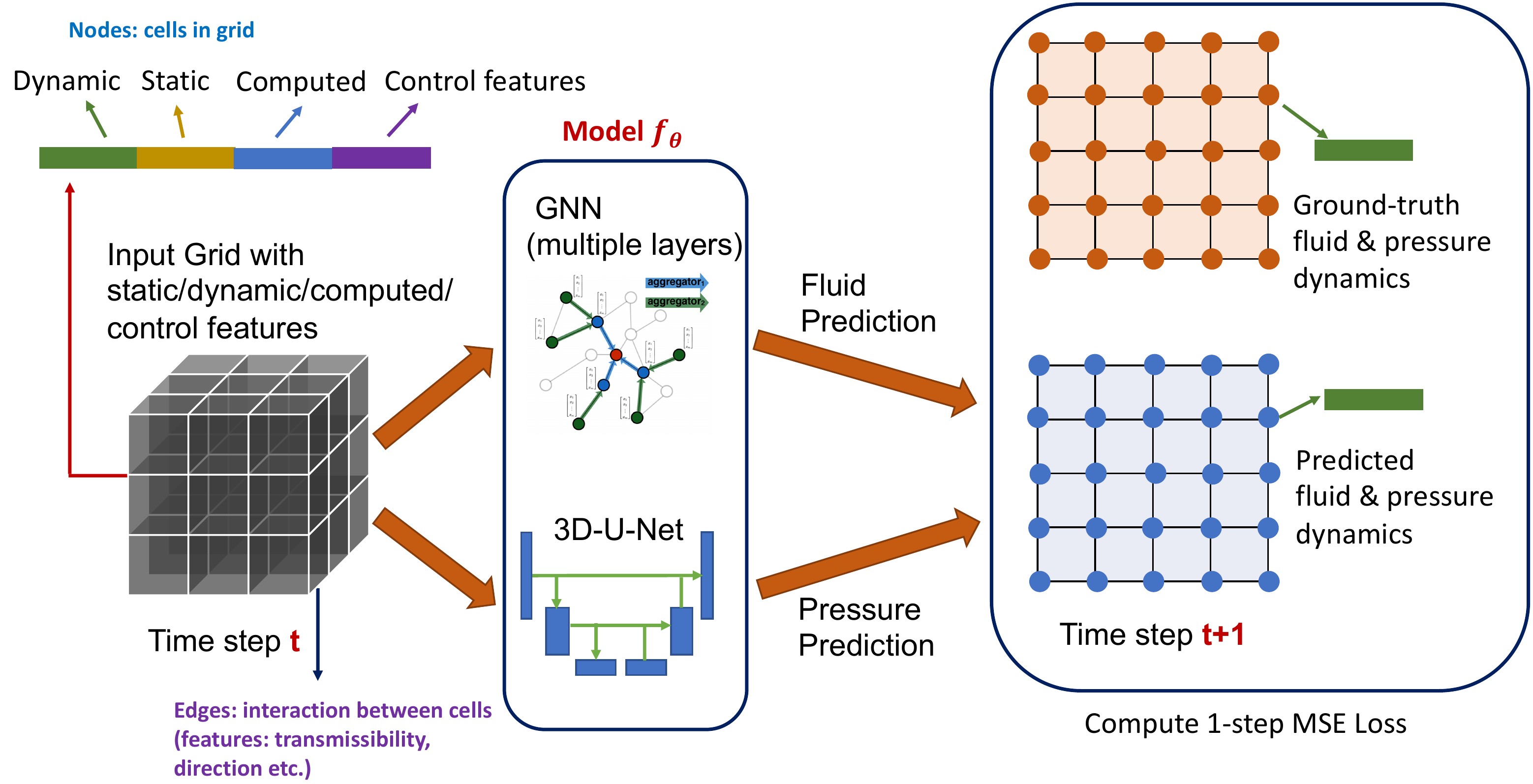}
    \caption{Overview of our HGNS architecture and 1-step loss computation. HGNS consists of a Subsurface Graph Neural Network (SGNN) to model the fluid dynamics, and a 3D-U-Net to model the pressure dynamics. The input grid on the left is treated as a graph by modeling each cell as a node, and connecting the adjacent cells via edges.}
    \label{fig:hgns}
\end{figure*}

The goal of simulations is to take as input static properties of the rock, initial states of quantities such as oil and water, and external control variables such as injection of water, and then predict the evolution of pressure and saturation of fluids over time (Fig.~\ref{fig:subsurface}). Two key challenges need to be addressed for practical, large-scale applications. First, subsurface flow is a complicated multi-scale problem. On the smaller spatial scale, it needs to model multiphase fluids flow through the complex subsurface structures: between neighboring cells (a cell is a discretization of space containing porous media through which fluid can flow), with various well configurations (wells can inject or produce fluids externally), and in presence of flow barriers (e.g., fault planes that can divert or prevent fluid paths). On a larger spatial scale, the flow of fluids is driven by convection forces and pressure gradient, whereas the dynamics of reservoir pressure is governed by global pore fluid distribution and reaches the equilibrium much faster. Therefore, 
we need to model both small and large spatial scales. Second, 
the models often needs to scale to extremely large grids.
For example, a standard industry problem typically
consists of millions or tens of millions of cells. 
Field development applications, such as well configuration and completion for optimal dynamic performance and improved sweep, could benefit significantly from 
subsurface simulators that can perform fast inference and scale to large grids.

Standard subsurface simulators employ domain-specific implicit partial differential equation (PDE) solvers.
For large grids with tens of millions of cells, they still need to solve an equation involving the full grid, which may be computationally exhaustive and requires substantial inference time. 
Recently, data-driven surrogate models provide a promising complementary approach~\cite{sanchez2020learning,mrowca2018flexible}. They learn directly from data and may alleviate years of engineering development. Moreover, prior works in other domains show that the models can learn forward evolution
\cite{sanchez2020learning} and can have larger spatial and temporal intervals~\cite{kochkov2021machine,um2020solver}.
However, these models are insufficient to address the two challenges of subsurface simulation.
They are not able to model the multi-scale dynamics, 
because they utilize Gaussian process~\cite{hamdi2017gaussian}, polynomial chaos~\cite{bazargan2015surrogate}, feed-forward neural network~\cite{costa2014application}, convolutional neural networks~\cite{zhu2018bayesian}, or recurrent R-U-Net~\cite{tang2020deep}, which cannot simultaneously model lower-level interactions between neighboring cells as well as global dynamics such as pressure. Additionally, they do not scale to large-scale simulations as they have only been applied to 2D grids with up to 10k vertices (while a practical simulation requires 3D grids), and up to 20k nodes for machine-learning-based surrogate models developed in other domains, e.g. GNS \cite{sanchez2020learning}, which is two to three orders of magnitude smaller than needed for standard industry applications. 

\vspace{1.7mm}
\xhdr{Present work} Here we introduce a Hybrid Graph Network Simulator (HGNS) for subsurface simulation, which addresses the above two challenges. HGNS consists of a Subsurface Graph Neural Network (SGNN) designed to model the fluid flow through the complicated subsurface structure, and a 3D-U-Net \cite{cciccek20163d} to model the more global dynamics of pressure. SGNN and 3D-U-Net cooperatively learn to evolve the subsurface dynamics. For large grids, which make training especially hard (exceeding GPU memory, training may take weeks), we developed a sector-based training. It allows training on grids of up to 1.1 million of cells,
two orders of magnitude larger than previous machine-learning-based surrogate models. The ability to deal with models of this size makes HGNS the first fully machine-learning-based subsurface model applied to realistic 3D scenarios. 

We use an industry-standard subsurface flow dataset to evaluate our model's generalization capabilities in a challenging setting where the initial conditions, static properties and well locations are different than that of training, and compare its performance with strong baselines. We show that HGNS produces more accurate long-term evolution, and outperforms other learning-based models by reducing long-term prediction errors by up to 21\%.
Moreover, our model running on a single GPU is between 2 to 18 times faster than a standard subsurface solver using 8 CPUs. 

\section{Related Work}

There has been extensive research for developing surrogate models for subsurface flow. One line of research focuses on physics-based methods, for example coarse-grid modeling, reduced-physics modeling, or proper orthogonal decomposition (POD)-based reduced-order modeling (ROM)~\cite{van2006reduced,cardoso2009development,he2014reduced,yang2016fast,jin2018reduced,he2013reduced,xiao2019non}. These methods typically simplify aspects of the problem to make it tractable. For example, POD-based ROM can be thought of as using a linear encoder to map the full trajectory into a low-dimensional space. In addition, Fraces et al.~\cite{fraces2020physics} use physics-informed neural networks (PINN) and apply transfer learning and generative methods to solve an inference problem for 2-phase immiscible transport. While Cai et al.~\cite{cai2022physics} and references therein provide a review of PINN for solving inverse problems in fluid mechanics, Fuks and Tchelepi~\cite{fuks2020limitations} demonstrate that physics-informed ML approaches fail to approximate the fluid-flow dynamics governed by nonlinear PDEs in the presence of sharp variations of saturation and propose the solution by adding a small amount of diffusion to the conservation equation.

\vspace{1.5mm}
\xhdr{Data-driven surrogate models} Data-driven surrogate modeling relies on the data to learn evolution models. Hamdi et al.~\cite{hamdi2017gaussian} utilize Gaussian process to model a 20-parameter unconventional-gas reservoir system. Bazargan et al.~\cite{bazargan2015surrogate} employ polynomial chaos to model a 40-parameter 2D fluvial channelized reservoir undergoing waterflooding. Apart from the above, feed-forward neural networks~\cite{costa2014application}, convolutional neural networks (CNN)~\cite{zhu2018bayesian}, recurrent R-U-Net~\cite{tang2020deep} are also used to learn subsurface surrogate models. However, the models above are insufficient when it comes to modeling how the fluid flows through the complicated  subsurface structures. These physics-based methods are not able to generalize to trajectories that are not sufficiently close to that of training~\cite{tang2020deep}, and are insufficient to model the nonlinear multiscale behavior of the subsurface flow. For example, it is hard for convolution-based models (e.g. CNN, U-Net) to model the fluid flow between normal cells, well perforations, and faults where fluid cannot flow through.  Our HGNS utilizes Graph Neural Networks (GNN) to model the fluid flow, whose relation-based prior can naturally capture the complicated interaction among the subsurface structures. For example, faults can be modeled by cutting the edges on a fault plane for the GNN, so by design the cells on the two sides cannot interact why CNN may need much data to learn that behavior. Moreover, prior works in data-driven subsurface modeling are limited to 2D subsurface simulations, with up to 10k cells. In comparison, our HGNS is the first learned surrogate model applied to 3D subsurface, and can operate on data two orders of magnitude larger than prior applications. A concurrent work \cite{maucec2022geodin} applies interaction network \cite{battaglia2016interaction} to model the subsurface fluid dynamics with up to 1 million cells, but requires the solver to provide the ground-truth for pressure.

\vspace{1.5mm}
\xhdr{Generic learned surrogate models} Outside of subsurface simulation, there has been active work to develop generic learned surrogate models. Our work builds upon prior work of Graph Network Simulators (GNS) \cite{sanchez2020learning}, a state-of-the-art GNN-based model for particle-based simulation. Compared to GNS, our Hybrid Graph Network Simulator (GHNS) is a hybrid architecture that only uses GNN to model the fluid part. Furthermore, we use a different method to construct node and edge features, compute the message, and incorporate the domain knowledge of subsurface simulation. Furthermore, our sector-based training allows HGNS to learn on datasets with at least two orders of magnitude more nodes than GNS, and our multi-step objective allows better long-term prediction.

\section{Preliminaries}

We consider the problem of subsurface simulation of oil-water flow, which models how the pressure and saturation of fluids evolve over time, given initial states, static properties of the rock, and external control variables such as injection of water. Here we present a simplified Partial Differential Equation (PDE) for the system:
\begin{equation}
\label{eq:pde}
\frac{\partial (\phi \rho_j S_j)}{\partial t} = \nabla\cdot(\frac{\rho_j}{\mu_j} k_{rj}(S_j) \mathbf{k}\nabla P)+q_j 
\end{equation}
Here $j=w,o$ denotes different components/phases, with $w$ for water and $o$ for oil.  $S_j\in[0,1]$ is the saturation for phase $j$ (saturation is the ratio between the present volume of component $j$ and the pore volume $V$ the rock can hold at a location), and $P$ is the pressure.  $\rho_j$ is phase density, $\mu_j$ is the phase viscosity, and $\phi$ is the rock porosity.  $k_{rj}(S_j)$ is the relative permeability that is a function of the saturation $S_j$, usually obtained via laboratory measurements. $\mathbf{k}$ is the absolute permeability tensor. $q_j$ is the source/sink term, which corresponds to the injecting or producing of component $j$ at the well location. In this equation, the saturation $S_j$ and pressure $P$ are dynamic variables that vary with time and space. $\phi$, $\rho_j$, $\mu_j$, $\mathbf{k}$ are static variables that are constant in time but typically vary in space. The injection of water is externally controlled, and the production of water/oil $q_j$ at producer wells is also a dynamic variable. We can intuitively understand this equation as follows: the change of water/oil saturation $S_j$ is due to two terms: the divergence of the flux $\Phi_j=-\frac{\rho_j}{\mu_j}k_{rj}(S_j)\mathbf{k}\nabla P$ and a source/sink $q_j$ term. The flux $\Phi_j$ is driven by the pressure gradient $\nabla P$, where fluids flow from places with higher pressure to those with lower pressure. Note that the  coefficient $k_{rj}(S_j)$ depends on the dynamic variable $S_j$, making the dynamics nonlinear.

Eq.~(\ref{eq:pde}) is a simplified model. The problem we consider here is much more complicated and high-dimensional. Current (non-machine learning) approaches use implicit methods to evolve the system, which involves solving a system of equations containing up to tens of millions of equations (each cell of the discretized grid contributes one equation). Solving such a system can be slow even after linearization. Moreover, in reality there are complex well configurations (vertical wells, slanted wells, horizontal wells), faults planes in rock that reduce, divert or prevent fluid flow. This in addition adds significant complexity to modeling and evolving the system in an accurate way.
In contrast, machine learning represents an attractive approach to alleviate these issues and develop faster, more scalable and accurate simulators of such complex phenomena.

\begin{figure*}[!ht]
    \centering
      \includegraphics[width=0.65\linewidth]{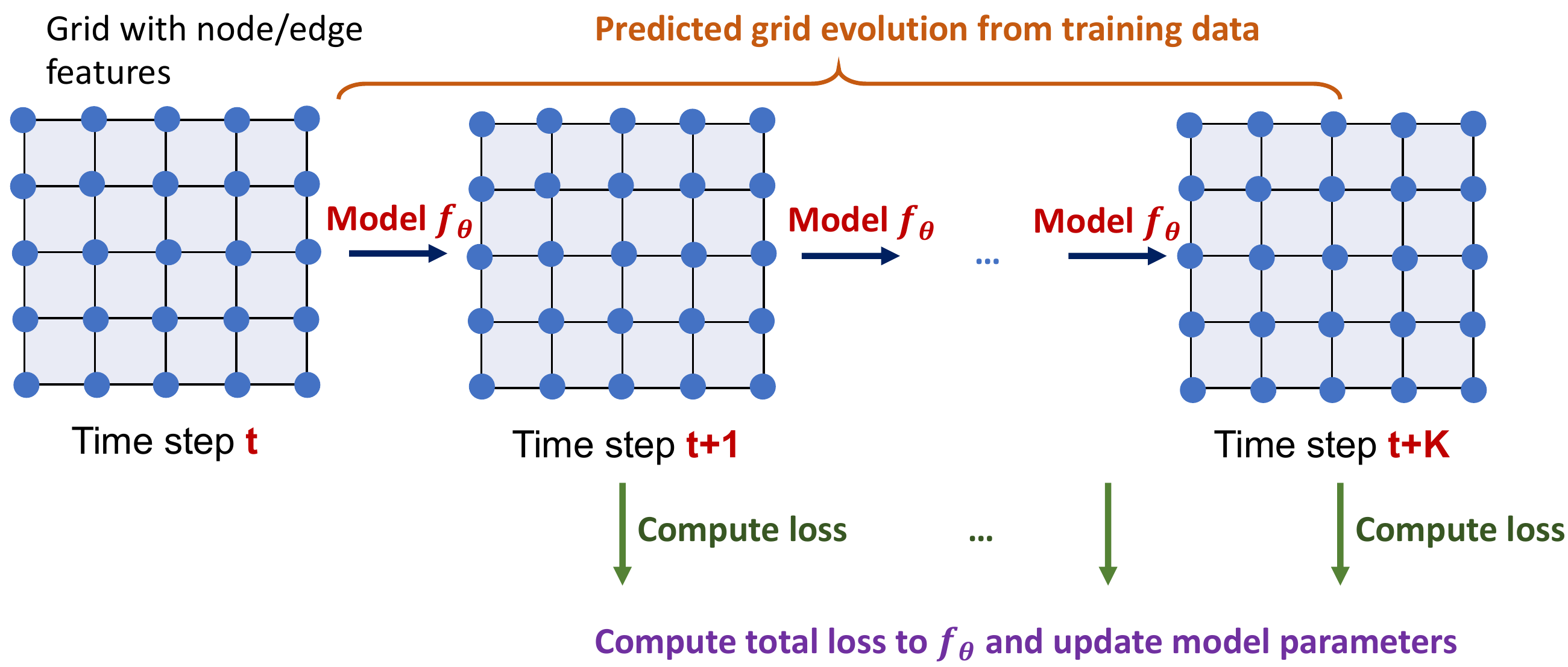}
    \caption{Multi-step rollout during training. The surrogate model is applied autoregressively to predict the future states, using its prediction of dynamic variables as the next prediction's input. The loss consists of error on each time step, and during training, the gradient can pass through the full rollout. This multi-step loss enables better long-term prediction.}
    \label{fig:multistep}
\end{figure*}

\section{Method}
In this section, we introduce our Hybrid Graph Network Simulator (HGNS) architecture (Fig.~\ref{fig:hgns}) and its learning method for subsurface simulation, addressing the two challenges raised in the introduction.

We formalize the problem of learning subsurface simulation as follows. A learnable simulator $f_\theta$ takes as input the dynamic variables $X^t\in \mathbb{R}^{N\times d_X}$ at time $t$, static variables $Q\in \mathbb{R}^{N\times d_Q}$, computed variables $R(X^t)\in \mathbb{R}^{N\times d_R}$, and control variables $U^t\in \mathbb{R}^{N\times d_U}$, all defined on the cells of a grid ($N$ is the number of cells), and predicts the dynamic variable $\hat{X}^{t+1}$ at the next time step:

\begin{equation}
\label{eq:prediction_1_step}
\hat{X}^{t+1}=f_\theta(X^t, Q, R(X^t), U^t)
\end{equation}
Here $d_X,d_Q,d_R,d_U$ are the number of features for each variable. The dynamic variable $X^t=(S_w^t, S_o^t, P^t)$ consists of water saturation $S_w^t$, oil saturation $S_o^t$ and pressure $P^t$. The static variable $Q$ consists of time-constant variables on each cell such as cell porosity, absolute permeability in $x, y, z$ directions, cell depth, and pore volume. The computed variable $R(X^t)$ consists of features that we opt to compute to facilitate the learning, for example the relative permeability $k_{rj}=k_{rj}(S_j)$. The control variable $U^t$ may include the injection of water at each time step. During inference, we can apply Eq. (\ref{eq:prediction_1_step}) in an autoregressive way to predict the long-term future, using the previous prediction as input, i.e.

\begin{equation}
\label{eq:prediction_multi_step}
\hat{X}^{t+k+1}=f_\theta(\hat{X}^{t+k}, Q, R(\hat{X}^{t+k}), U^{t+k}), k=0,1,...K
\end{equation}
where we only provide $\hat{X}^t=X^t$ as initial state and also provide $U^{t+k},k=1,2,...K$ as external control at each time step.
The goal of learning is to learn the parameter $\theta$ of $f_\theta$ such that the prediction $\hat{X}^{t+k}$ is as near the ground-truth $X^{t+k}$ as possible, even if $k$ is large (a long-term prediction).

\subsection{HGNS Architecture}

To address the multi-scale behavior of the subsurface dynamics, we introduce a Subsurface Graph Neural Network (SGNN) to model the dynamics of fluids (water, oil) on a finer scale, and a 3D-U-Net~\cite{cciccek20163d} to model the more global dynamics of pressure. They combine to form the HGNS architecture. 

Concretely, our HGNS $f_\theta$ can be written as:

\begin{align}
\left\{
\begin{array}{ll}
  \hat{S}^{t+1}=g_\theta(X^t,Q,R(X^t),U^t)+S^t\\
  \hat{P}^{t+1}=h_\theta(X^t,Q,R(X^t),U^t)+P^t
 \end{array}
 \right.
\end{align}
Here $\hat{S}^t=(S_w^t,S_o^t)$ is the saturation for water and oil. $g_\theta$ is the SGNN model and $h_\theta$ is a 3D-U-Net \cite{cciccek20163d}. They both predict the \emph{change} of the dynamic variables rather than the value itself.
Further details about these two components and the computed features $R(X^t)$ are provided next.
In order to effectively model subsurface simulation through message passing, we also highlight the domain-specific designs of SGNN, such as the use of transmissibility for message computation, and augmentation of computed features as a prior.

\medskip

\xhdr{Subsurface Graph Neural Network (SGNN) for Fluid Saturation Prediction}
Our SGNN uses the encoder-processor-decoder architecture, similar to \cite{sanchez2020learning}. 

\paragraph{Encoder} The encoder embeds the input $(X^t,Q,R(X^t),U^t)$ into a latent graph $\G^{(0)}=(\mathcal{V}^{(0)},\mathcal{E}^{(0)})$, where $\mathcal{V}^{(0)}=\{v_i^{(0)}\}$ is the collection of input node features and $\mathcal{E}=\{e_{ij}^{(0)}\}$ is the collection of edge features.
Specifically, we regard each cell as a node, which is connected to its 6 neighbors (since we are modeling 3D grids) with bidirectional edges. Each node has complete features $(X^t,Q,R(X^t),U^t)$, and the edge features $e_{ij}^{(0)}$ consist of transmissibility (capturing how easy it is for the fluid to flow between neighboring cells) and a one-hot indicator of the direction of the edge (since we consider gravity, the edge pointing up, down and horizontally should be treated differently). After obtaining $\G^{(0)}$, the encoder has a Multilayer Perceptron (MLP) that encodes the node features $\mathcal{V}$ into some latent embedding. We will detail the specific features encoded in Appendix \ref{app:features}, as well as Appendix A.2.

\paragraph{Processor} The processor is a stack of $M$ graph neural network layers, each one performing one round of message passing that mimics the flow of fluid between neighboring cells. Specifically, given the latent graph $\G^{(m)}=(\mathcal{V}^{(m)},\mathcal{E}^{(m)})$ outputted from the previous layer, it first computes the message on the edges based on neighboring nodes:
\begin{equation}
\label{eq:edge_update}
e_{ij}^{(m)}=MLP_e([v_i^{(m)}, v_i^{(m)} - v_j^{(m)}, e^{(0)}_{ij}])
\end{equation}
where $e_{ij}^{(m)}$ is the message from node $i$ to node $j$ on the ${m-1}^\text{th}$ layer, ``$[]$'' is concatenation on the feature dimension. The above Eq. (\ref{eq:edge_update}) utilizes the input edge feature $e_{ij}^{(0)}$, the source node $i$'s feature, and the difference between source node $i$ and target node $j$'s features, to compute the message using an MLP$_e$. After computing the message, each node will aggregate the messages sent to it by summation, obtains $e_j^{(m)'}$, and then performs node update:

\begin{equation}
\label{eq:node_update}
v^{(m+1)}_j = MLP_v([v_j^{(m)}, e_j^{(m)'}])
\end{equation}

Then we perform group normalization \cite{wu2018group} with 2 groups before the next layer. The value of our specific hyperparameter is given in Appendix \ref{app:hyperparameter}.

\paragraph{Decoder} We use an MLP that maps the output of the processor back to the predicted dynamic variables at the next time step.

Overall, our SGNN models the complex subsurface flow by encoding the properties and dynamics of each cell and cell-cell relation into node and edge features, then uses an expressive edge-level MLP$_e$ to compute the interaction between neighboring cells, and a node-level MLP$_v$ to update the state of the cells. In this way, it can model subsurface flow through complex subsurface structures. 

\medskip
\xhdr{3D-U-Net for Pressure Prediction}
\label{sec:3d_unet} The subsurface pressure dynamics has faster equilibrium time than fluid, and therefore its evolution is more global within the time scale (months) of fluid. Therefore, we utilize 3D-U-Net \cite{cciccek20163d} to model the dynamics of pressure, whose hierarchical structure can capture this more global dynamics.
We modify 3D-U-Net's order of operation such that for each convolution layer, it first performs 3D-convolution, then applies activation (ReLU) followed by group normalization. We make this modification since we observe that this improves performance compared to the default order of Conv $\to$ BN $\to$ ReLu.

\medskip
\xhdr{Computed Features} As shown in Eq. (\ref{eq:pde}), the coefficient of the flux $\Phi_j=-\frac{\rho_j}{\mu_j}k_{rj}(S_j)\mathbf{k}\nabla P$ depends on the relative permeability $k_{rj}(S_j), j=w,o$, which is a function of the dynamic variable $S_j$ and the function is provided via a table of laboratory measurements. Thus, to facilitate neural network's modeling, we add $[k_{rw},k_{ro}]=\left[k_{rw}(S_w),k_{ro}(S_o)\right]$ as computed features, using linear interpolation to compute the value between the measured points. Furthermore, we add spatial gradient of the dynamic variable $\nabla X$ as another part of computed features, to allow the neural network better to utilize a cell's neighboring feature to predict its future (e.g. as shown in Eq. \ref{eq:pde} the flux $\Phi_j$ is proportional to the spatial gradient of pressure $\nabla P$). Overall, we have that the computed feature is given by 
\begin{equation}
R(X^t)=[k_{rw}(S_w^t),k_{ro}(S_o^t), \nabla X^t]
\end{equation}

\subsection{Training}

The goal of the training is for the model to have small long-term prediction error for grids consisting of millions or tens of millions of cells. To address the challenge of the large grid size, we introduce sector-based training. For more accurate long-term prediction, we develop the technique of multi-step rollout during training.

\medskip
\xhdr{Sector-based Training} Suppose that the grid has 10 million cells, training such large size will certainly exceed the memory limit of a single GPU, which typically has up to 32GB of memory. 

To address this problem, we note that both the SGNN and the 3D-U-Net are ``local'' models, meaning that to predict the cell's state at $t+1$, we only need a cell's neighbors up to a certain distance away, instead of the full grid. Based on this observation, we partition the full grid into ``sectors'', where each sector is a cube of cells (e.g. $40\times 40 \times 40$), with strides (e.g. 20) in each direction to obtain all sectors. Then during training, the model only needs to make predictions on the sectors instead of the full grid. In this way, we can train with a full grid of arbitrary size, because we can always partition the full grid into sectors with constant size to fit into GPU's memory.

To properly perform sector-based training, a few things need to be taken into account, and below is our design to address them:

\begin{itemize}
\item \textbf{Boundary of the sectors:} The cells near the boundaries may suffer from artificial boundary effect, where they lack neighboring cells in certain directions to properly perform prediction. The total loss will be biased if we include the loss on those cells. To address this, we use a sector-specific mask which masks out the cells up to a certain distance from the sector's boundary (the cells on the real boundary of the full grid are not masked out). The cells masked out will not contribute to the loss during training.
\item \textbf{Sector stride:} Because we neglect the loss on the sector's boundary, if the stride is the same as the sector's size, some cells will never contribute to the loss. To address this, we make the stride smaller than the sector's size, making sure that all cells must contribute to the loss computation in at least one sector.

\item \textbf{Mixing sectors:} The governing equation for the dynamics is the same regardless of where the sector is, what time step the evolution is at, or which trajectory the grid corresponds to. Therefore, it is beneficial to mix sectors randomly across all places, time steps and trajectories into minibatches during training. This way, we can reduce the correlation between examples and help the network learn the dynamics that can be applied in a more general way.
\end{itemize}

The sector-based training also enables multi-GPU training. If training on a single GPU, a typical training with 40 epochs will take approximately 1-2 weeks, since each epoch typically needs to go through 1 to 10 billion cells due to multiple trajectories and up to hundreds of time steps per trajectory. Sector-based training allows us to assign sectors into multiple GPUs, where the GPUs accumulate gradients to a central GPU. With $n$ GPUs, the time typically reduces by $n/2$ to $n$ fold with some offset. This reduces our training time to around 2 days.

In summary, sector-based training makes training such large grid possible, allowing us to train with grids that have at least 2 orders of magnitude more cells than prior learning-based subsurface models. It also makes training with multi-GPU possible and reduces training time significantly.

\medskip
\xhdr{Multi-step Rollout During Training} A standard learning objective for learned simulation is to minimize the following Mean Squared Error (MSE) on the 1-step prediction:

\begin{equation}
L= \mathbb{E}_t\left[\ell(f_\theta(X^t,Q,R(X),U), X^{t+1})\right]
\end{equation}
where $\ell(\hat{X}^{t+1},X^{t+1})=(\hat{X}^{t+1}-X^{t+1})^2$.  However, we find that even with a very small 1-step training loss, the multi-step rollout (Eq. \ref{eq:prediction_multi_step}) can have a large error due to error accumulation. To improve long-term prediction, we develop multi-step rollout during training. It performs multiple steps of rollout, and the loss is given by:

\vspace{-3mm}
\begin{align}
\left\{\begin{array}{ll}
L=\mathbb{E}_t\left[\sum_{k=1}^{K}\lambda_k\cdot \ell\left(f_\theta(\hat{X}^{t+k},Q,R(\hat{X}^{t+k}),U^{t+k}), X^{t+k+1}\right)\right]\\
\hat{X}^{t+k}=f_\theta(\hat{X}^{t+k-1}, Q, R(\hat{X}^{t+k-1}), U^{t+k-1}),k=1,2,...K
\end{array}
\right.
\end{align}
Here $\hat{X}^t:=X^t$ is the initial state for the rollout. Fig. \ref{fig:multistep} illustrates the multi-step loss. It uses the model $f_\theta$ to perform rollout autoregressively, using the predicted $\hat{X}^{t+k}$ as the next state's input, and the loss is a weighted sum of loss on the autoregressive predictions $\hat{X}^{t+1},...,\hat{X}^{t+K+1}$ for all rollout steps, weighted by weights $\lambda_k$. During training, the backpropagation can pass through the full rollout steps, so that the model is \emph{directly} trained to minimize long-term prediction error. In practice, there is a tradeoff between computation and accuracy. The larger the total rollout steps $K$ in training, the better potential accuracy it can achieve, but the more compute and memory it requires (scales linearly with $K$). We find that using $K=4$ strikes a good balance, which uses reasonable compute, achieves far better accuracy than 1-step loss, and there is minimal gain to further increase $K$. In our experiments, we use $K=4$ and $(\lambda_1, \lambda_2, \lambda_3, \lambda_4)=(1,0.1,0.1,0.1)$. The weights for the steps $>1$ are smaller, since at the beginning of training when the model is inaccurate, the multi-step loss is much worse. Having a larger weight on those steps would make the model not able to learn anything. Having 1-step loss dominates as we are using helps the model to find a good minimum of the loss landscape first, and can further improve by the multi-step part.

\begin{figure}[t]
    \centering
      \includegraphics[width=0.7\linewidth]{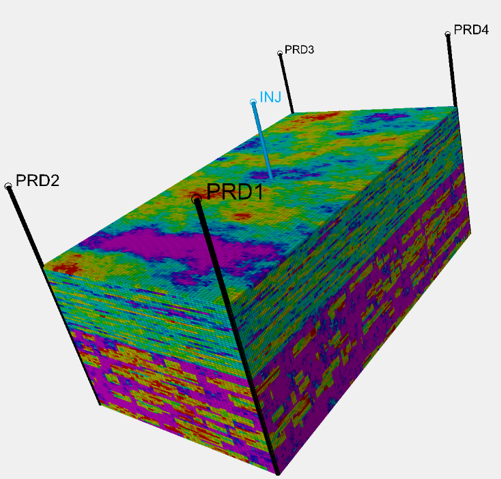}
    \caption{An example configuration of the dataset we use (SPE-10 model). The color shows the absolute permeability in the x direction ($k_x$), with blue, purple and red representing values from low to high. There is one injector (INJ) in the middle and 4 producers (PRD) at the four corners.}
    \label{fig:dataset_vis}
    \vspace{-4mm}
\end{figure}

\vspace{-1mm}
\section{Experiments}

To evaluate our HGNS model we examine: (1) Does our HGNS model achieve better long-term prediction performance than other strong baselines? And, (2) How do different components of HGNS, e.g. hybrid design, multi-step rollout, contribute to predictive performance? We evaluate the models in a challenging setting in which the static properties, initial conditions and the well locations in testing are all different from that during training. In this way, we measure how the models are able to generalize to novel scenarios.

\vspace{1mm}
\xhdr{Dataset and training regime}
We evaluate our model on an industry-standard subsurface simulation dataset with single generated by a standard solver. It is a Single Porosity Single Permeability (SPSP) model \cite{spe10dataset}, which does not have fractures. It consists of 20 trajectories, each trajectory has 61 time steps (the time interval between neighboring steps is 1-month), and the full grid has ~1.1 million cells (thus each epoch needs to go through 1.37 billion cells). These trajectories differ in their static properties, initial states and well locations. See Fig. \ref{fig:dataset_vis} for an example configuration of the input grid. We randomly choose 16 trajectories for training, and the other 4 trajectories for testing. Since different trajectories have drastically different static properties, initial states and well locations, the performance at the test set evaluates how the models are able to generalize to novel subsurface structures. In the testing, we provide the models the initial state at $t=3$ (allowing for 3 steps of initial stabilization), and autoregressively roll out the models for 10 and 20 steps, and compute the Mean Absolute Error (MAE) between the model's prediction and the ground-truth. In the experiment, we use volume of water $V_w=S_w\cdot V$ and oil $V_o=S_o\cdot V$ instead of saturation $S_w, S_o$ as dynamic variable (here $V$ is the pore volume that is a static property denoting the total volume of fluid a cell can hold, $V_w+V_o=V$), to allow the models to better utilize mass conservation.

\vspace{1mm}
\xhdr{Baselines}
We compare our HGNS model, trained with 4-step rollout, with two other strong baselines, 3D-U-Net \cite{cciccek20163d} and standard CNN, each trained with 4-step rollout, to test how the architectures influence the performance. In addition, we also compare with a ``predict no change'' baseline, in which the model simply copies the previous time step as its prediction. In this way, it gives a baseline scale of error. Table \ref{tab:exp_res1} shows the result.
The ground-truth pressure is typically on the level of 6000-10000 psi, and the water and oil volume is typically on the level of 10-20 barrels. Typically, a 100 psi error on the pressure and 1 barrel error on the water/oil on a cell in a long-term prediction is deemed as acceptable.
Note that the compared models already use our encoding of computed features,
sector-based training,
and multi-step rollout during training for some models; three techniques that enable training of such large datasets and deliver better performance.

\begin{table*}[h!]
\centering
\caption{Mean Absolute Error (MAE) of our HGNS and other baseline models on the test set for pressure, water and oil prediction, after 10-step (10 months) and 20-step rollout. HGNS outperforms other strong baselines by an pressure error reduction of 5.5\% and 6.7\% for 10-step and 20-step, water error reduction of 21\% and 8.4\% for 10-step and 20-step, and oil error reduction of 21\% and 8.3\% for 10-step and 20-step, compared with the best performing scenario in other models.
}
\begin{tabular}{@{}c|ccc@{}|ccc@{}}
\Xhline{2\arrayrulewidth}
\multicolumn{1}{c|}{} & \multicolumn{3}{c|}{10-step prediction MAE} & \multicolumn{3}{c}{20-step prediction MAE}\\ 
\cline{2-7}
Model       & Pressure (psi) & \makecell{Water \\ (barrel)}  & \makecell{Oil\ \  \\ (barrel)\ \ } & Pressure (psi) & \makecell{Water \\ (barrel)}  & \makecell{Oil \\ (barrel)}\\ 
\hline
Predict no change         &     210.8    &         0.941         & 0.941  &      296.1    &         1.541        & 1.541 \\
CNN          & 77.9         & 0.628                 & 0.608 & 104.2        & 1.157                & 1.090 \\
3D-U-Net      & 94.6         & 0.361                 & 0.361  & 142.6         & 0.725                 & 0.724 \\
\hline
\textbf{HGNS (ours)} & \textbf{73.6}  & \textbf{0.286} & \textbf{0.286} & \textbf{97.2}  & \textbf{0.664} & \textbf{0.664} \\
\Xhline{2\arrayrulewidth}
\end{tabular}
\label{tab:exp_res1}
\end{table*}

\begin{table*}[h!]
\centering
\caption{Mean Absolute Error (MAE) of our HGNS and its ablations on the test set for pressure, water and oil prediction, after 10-step (10 months) and 20-step  rollout. Hybrid design and multi-step training of HGNS improve performance by an average error reduction of 15.7\% and 22.4\%, respectively.
}

\begin{tabular}{@{}c|ccc@{}|ccc@{}}
\Xhline{2\arrayrulewidth}
\multicolumn{1}{c|}{} & \multicolumn{3}{c|}{10-step prediction MAE} & \multicolumn{3}{c}{20-step prediction MAE}\\ 
\cline{2-7}
Model       & Pressure (psi) & \makecell{Water \\ (barrel)}  & \makecell{Oil \ \ \\ (barrel)\ \ } & Pressure (psi) & \makecell{Water \\ (barrel)}  & \makecell{Oil \\ (barrel)}\\ 
\hline
\textbf{HGNS (ours)} & 73.6  & \textbf{0.286} & \textbf{0.286} & \textbf{97.2}  & \textbf{0.664} & \textbf{0.664}  \\
\hline
HGNS without 3D-U-Net (only SGNN) & 74.8&    0.307  & 0.307  & 110.5  & 0.829 & 0.829 \\
HGNS without SGNN (only 3D-U-Net)     & 94.6         & 0.361                 & 0.361  & 142.6         & 0.725                 & 0.724 \\
HGNS with 1-step  & \textbf{47.3}    & 0.500 & 0.500    & 122.8        & 1.144      & 1.144 \\
\Xhline{2\arrayrulewidth}
\end{tabular}
\label{tab:exp_res2}
\end{table*}

\begin{figure*}[t]
\centering

\begin{subfigure}[b]{0.32\textwidth}
\centering
   \includegraphics[width=0.9\linewidth]{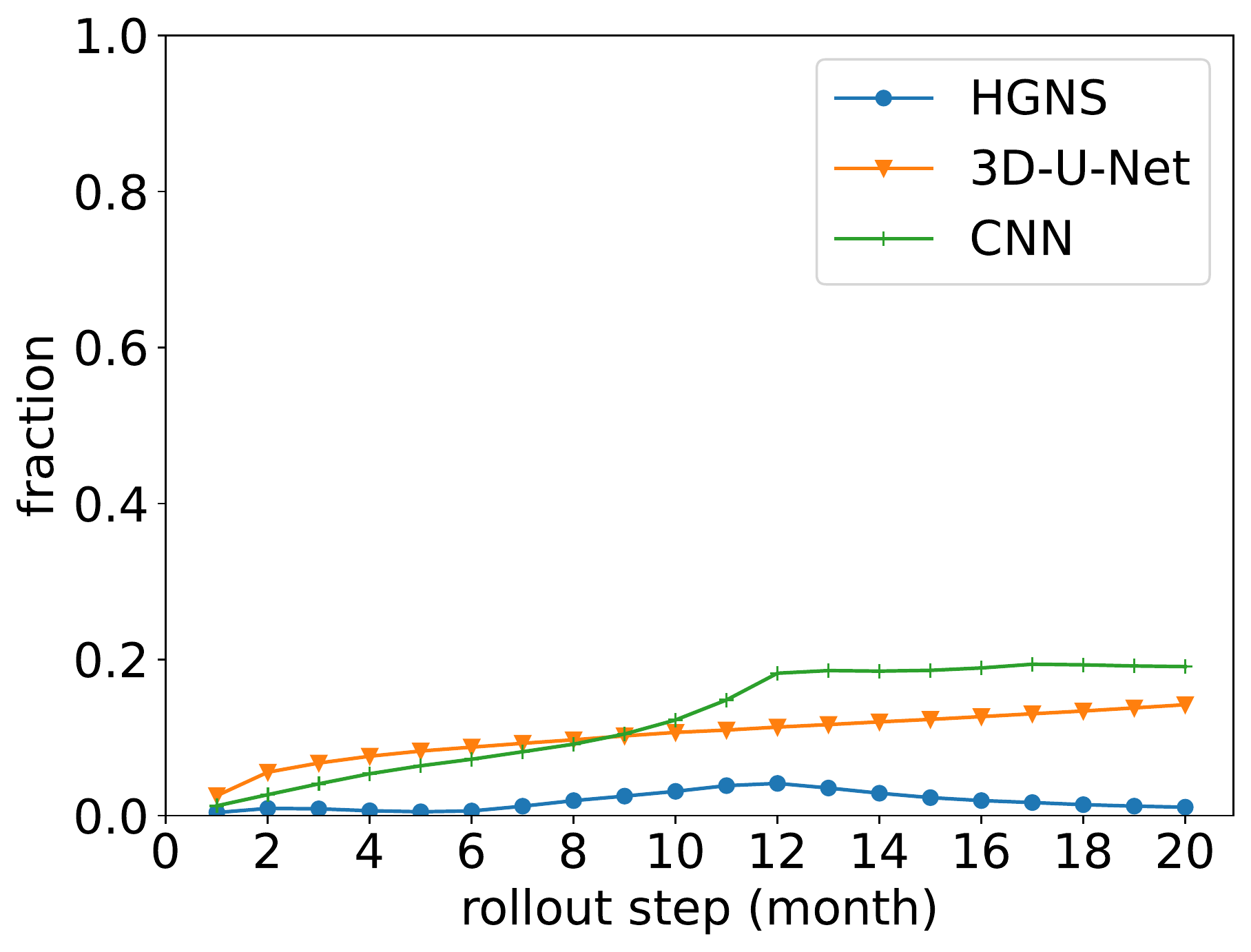}
   \caption{Fraction of cells whose absolute error of pressure is greater than 100 psi.}
   \label{fig:pressure_valid_fraction}
\end{subfigure}
~\ \ \ \ \ \ \ \ 
\begin{subfigure}[b]{0.32\textwidth}
\centering
   \includegraphics[width=0.9\linewidth]{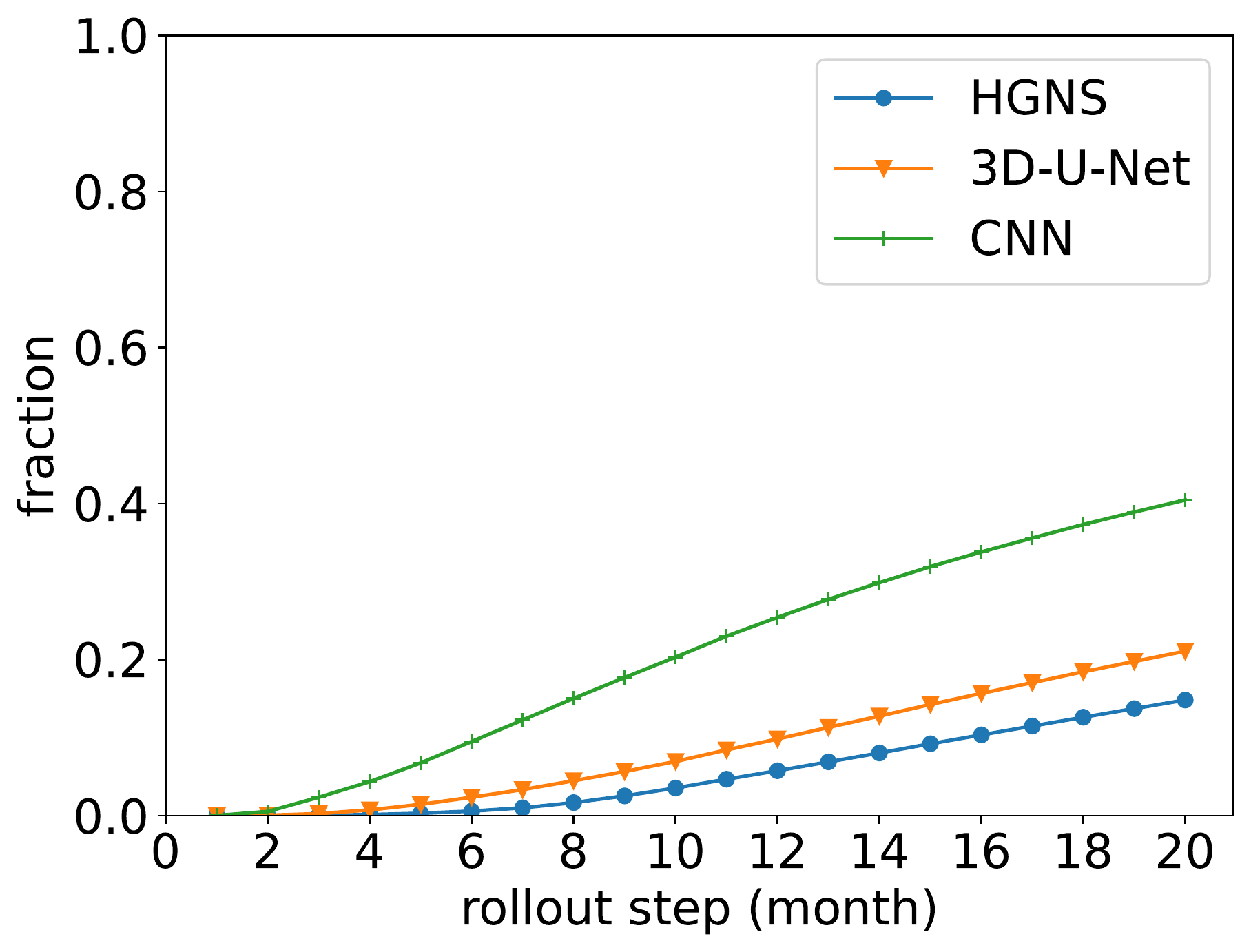}
   \caption{Fraction of cells whose absolute error of water volume is greater than 1 barrel.}
   \label{fig:water_valid_fraction} 
\end{subfigure}
~\ \ \ \ \ \ \ \ 
\begin{subfigure}[b]{0.32\textwidth}
\centering
   \includegraphics[width=0.9\linewidth]{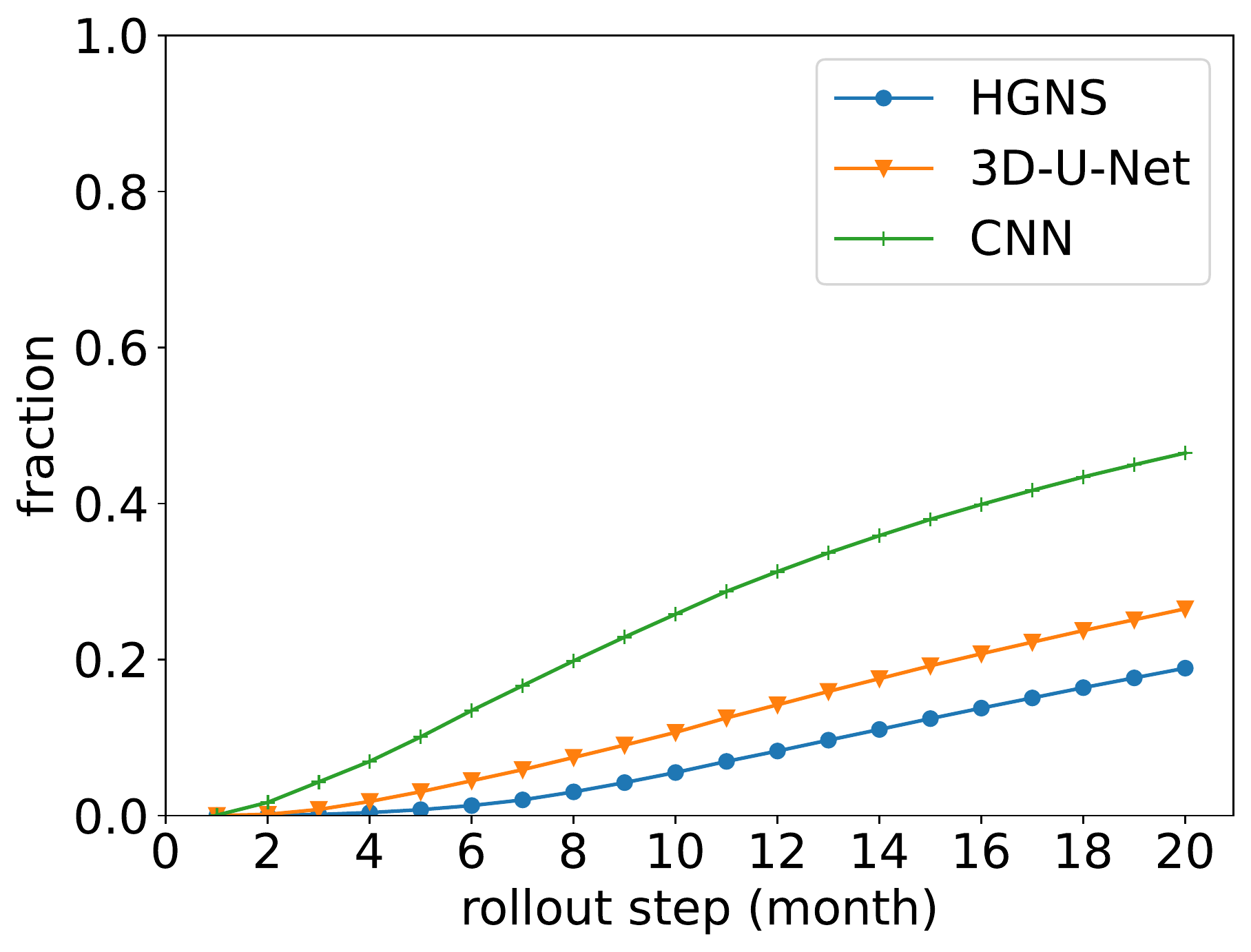}
   \caption{Fraction of cells whose absolute error of oil volume is greater than 1 barrel.}
   \label{fig:oil_valid_fraction} 
\end{subfigure}

\caption{Comparison of models during rollout on the fraction of cells whose error of (a) pressure (b) water volume (c) oil volume is above a given threshold. The lower the fraction, the better the prediction. HGNS outperforms 3D-U-Net and CNN in both scenarios, achieving a reduction of fraction of 71\% for pressure, 29.7\% for water volume, and 28.7\% for oil volume, compared with the best performing model.}
\label{fig:fraction}
\end{figure*}

\begin{figure*}
\centering
\begin{subfigure}[]{0.7\textwidth}
   \includegraphics[width=1\linewidth]{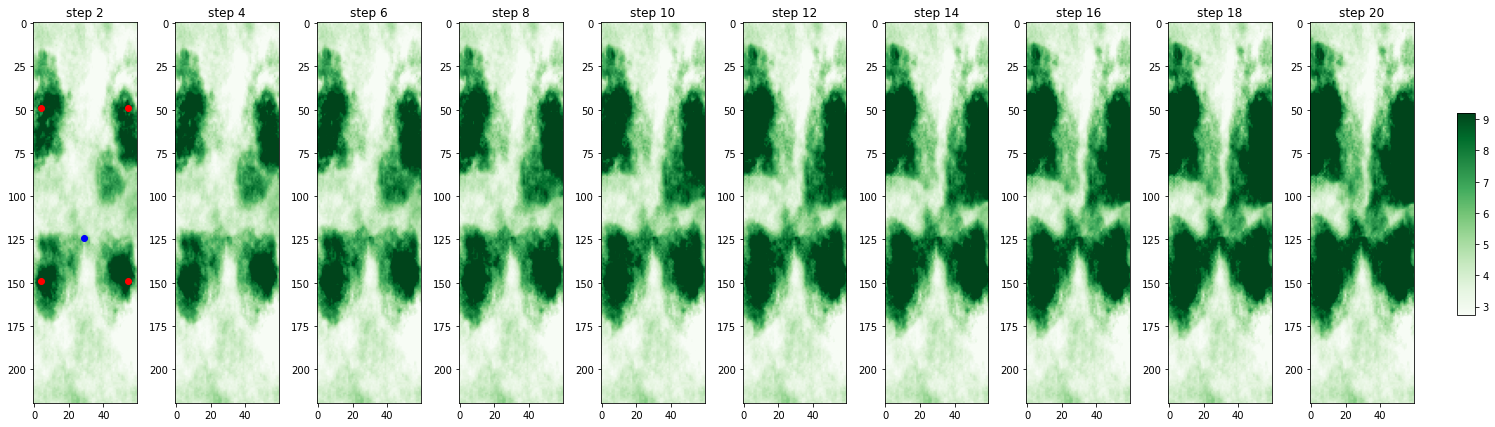}
   \caption{HGNS rollout of water volume (barrel) for 20 steps (months)}
   \label{fig:water_prediction_pred} 
\end{subfigure}

\begin{subfigure}[b]{0.7\textwidth}
   \includegraphics[width=1\linewidth]{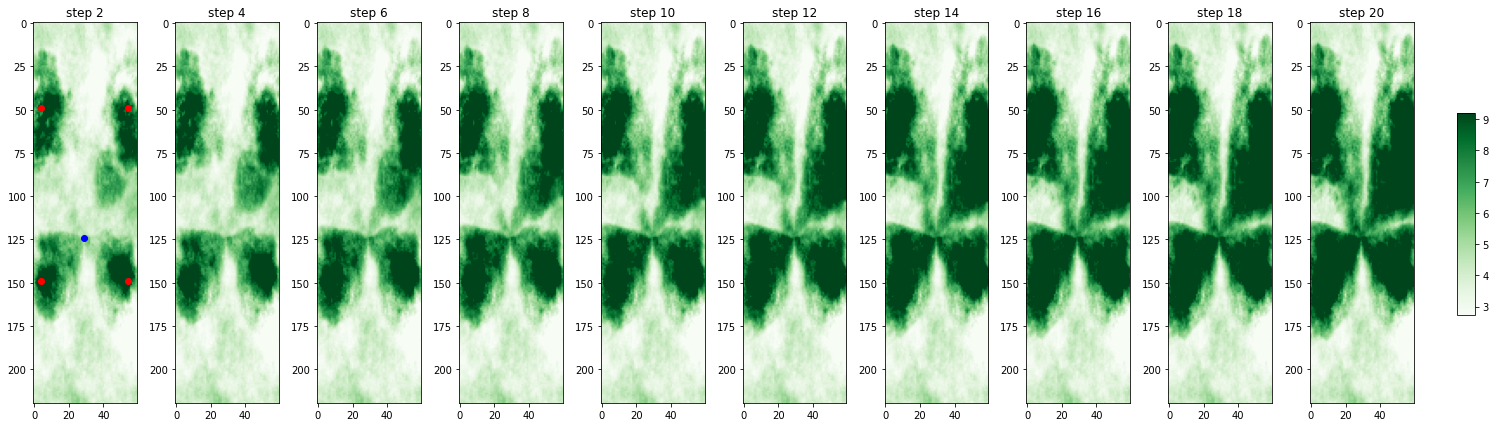}
   \caption{Ground-truth of water volume (barrel) for 20 steps}
   \label{fig:water_prediction_target}
\end{subfigure}

\begin{subfigure}[b]{0.7\textwidth}
   \includegraphics[width=1\linewidth]{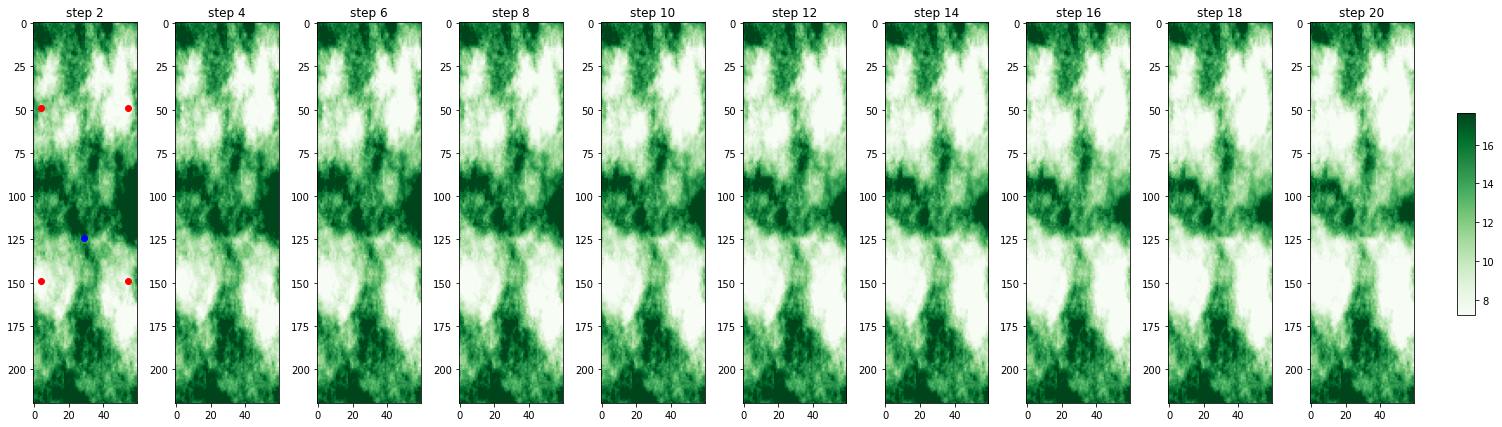}
   \caption{HGNS rollout of oil volume (barrel) for 20 steps}
   \label{fig:oil_prediction_pred} 
\end{subfigure}

\begin{subfigure}[b]{0.7\textwidth}
   \includegraphics[width=1\linewidth]{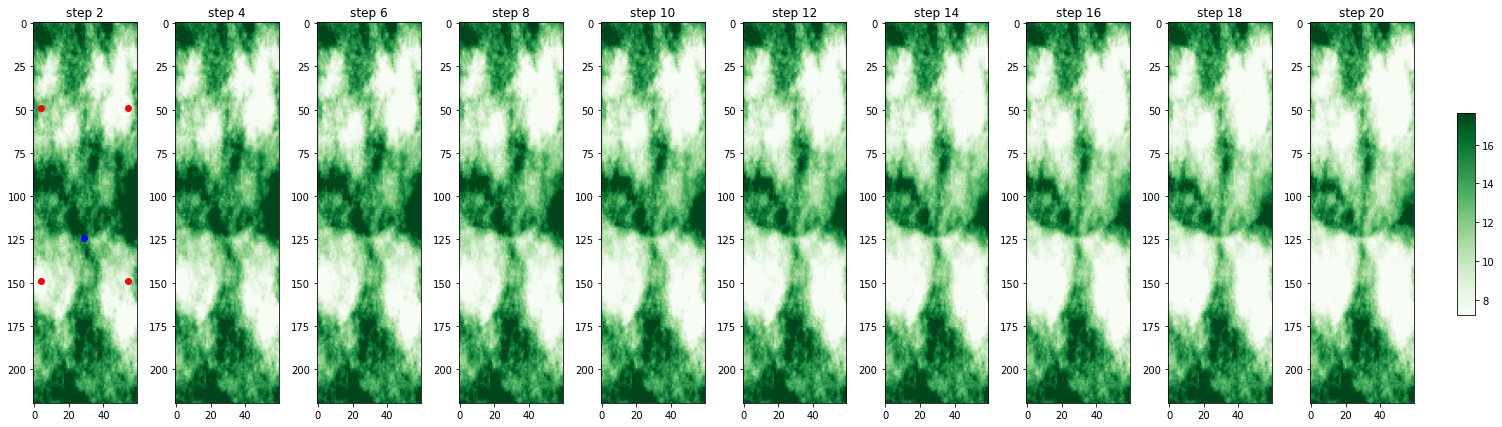}
   \caption{Ground-truth of oil volume (barrel) for 20 steps}
   \label{fig:oil_prediction_target}
\end{subfigure}

\caption{Comparison between (a) HGNS 20-step (20-month) rollout of water volume vs. (b) ground-truth water dynamics, and (c) HGNS 20-step rollout of oil volume vs. (d) ground-truth oil dynamics, on one of the trajectories, at a cross section of depth 20. Notice that HGNS reliably captures water  flow from 4 injectors (red dots) to the producer (blue dot at the middle), and the oil flow and drainage due to the producer.
}
\label{fig:vis_bwat}
\end{figure*}

\vspace{1mm}
\xhdr{Accuracy of 10- and 20-month predictions}
From Table \ref{tab:exp_res1}, we see that our HGNS outperforms other baselines by a wide margin, achieving 5.5\% and 6.7\% reduction of pressure error at 10-step and 20-step rollout, 21\% and 8.4\% reduction of water error at 10-step and 20-step, and 21\% and 8.3\% reduction of oil error\footnote{In this dataset, since we only have two phases of fluids (water and oil), their volume sum to the pore volume (a static variable): $V_w+V_o=V$. All models seem to learn this conservation law, resulting in similar errors between water and oil.} at 10-step and 20-step than the best-performing model. We also see that the error of HGNS is significantly less than the ``predict no change'' baseline, showing that it has learned non-trivial dynamics of the system.

Another important angle to evaluate the prediction is to compute the \emph{fraction} of cells whose prediction error is less than 100 psi for pressure and 1 barrel for water/oil volume. Fig. \ref{fig:fraction} shows the above fractions vs. rollout steps for HGNS and the two other compared models. We see that HGNS outperforms the other models consistently across all rollout steps, in fraction for pressure, water and oil. Moreover, even after 20 steps (months) rollout of HGNS, the fraction of cells whose pressure error is greater than 100 psi remains below 4.1\%, 71\% lower than the best performing model (3D-U-Net) whose maximum fraction is 14.2\%. Similarly, HGNS's fraction of cells whose water and oil volume error is below than 1 barrel remains below 14.8\% and 18.9\%, respectively, 29.7\% and 28.7\% lower than the best performing model. This shows that the prediction of HGNS is above standard for a majority of cells, and achieve significant improvement over strong baselines.

We visualize our HGNS's 20-step water and oil prediction and compare it with ground-truth, in one typical test dataset, as shown in Fig. \ref{fig:vis_bwat}.
Fig. \ref{fig:vis_bwat} (a) and (b) show that our model captures reliably the water flow from the injector (located at the corners and waist) to the middle producer. (c) and (d) show that our model captures the oil flow and drainage due to the producer. 

\vspace{1.7mm}
\xhdr{Ablations}
To evaluate how the hybrid design and multi-step training contribute to the improved performance, we compare HGNS with its ablations: one without 3D-U-Net (using SGNN to predict both the pressure and fluid), one without SGNN (using 3D-U-Net to predict pressure and fluid, same as in Table \ref{tab:exp_res1}), and one with only 1-step loss. The results are shown in Table \ref{tab:exp_res2}.

From Table \ref{tab:exp_res2}, we see that using only 3D-U-Net and SGNN to predict the full pressure and fluid results in worse performance than the hybrid design, confirming that our hybrid design better captures the global dynamics of pressure and more fine-grained dynamics of fluid flow. We also see that without 4-step loss, HGNS renders a much worse result. Even though HGNS 1-step has a better 10-step pressure error than HGNS, its longer-term pressure prediction (20-step) is worse, demonstrating that our multi-step training helps improving long-term prediction.

\xhdr{Runtime comparison} One advantage of our HGNS, compared with standard subsurface PDE solvers, is that it can perform explicit forward prediction to obtain the state at the next time step, while standard subsurface PDE solvers need implicit method to predict the next time step due to numerical stability issues. The implicit method requires solving an equation for the full grid, and the larger the grid, the slower it is to solve such systems of equations. On the dataset above with 1.1 million cells per trajectory, our HGNS took 20.7s to roll out 20-steps with an NVIDIA Quadro RTX 8000 GPU, compared to approximately 46s-370s (varying depending on the number of wells) required by the standard solver using 4 compute nodes, each with 2 CPUs Intel(R) Xeon(R) E5-2680 v3 2.50GHz, a 2 to 18-fold reduction in execution time. We expect that HGNS gains will be even more prominent with larger grid sizes (e.g. over 10 million).

\xhdr{Industry deployment} We are finishing up deploying and integrating our HGNS into an industry pipeline, for it to be used for speeding up the subsurface simulations and field development planning. We have addressed many additional engineering challenges during deployment, e.g. difference in the research and production environment, scaling up from the current 1.1 million cell model to significantly larger grid sizes, etc. The HGNS in deployment shares the same interface as the standard solver in the pipeline. During field development planning, HGNS will be used for fast rollout and finding candidate solutions to accelerate high-level well placement and operational decisions.

\vspace{-1mm}

\section{Conclusion}

In this work, we have introduced a Hybrid Graph Network Simulator (HGNS) for learning subsurface simulations. It employs a hybrid Subsurface Graph Neural Network (SGNN) to model the fluid flow through the complicated subsurface structures, and a 3D-U-Net to model the more global dynamics of pressure, addressing the challenge of multi-scale dynamics. We introduce sector-based training, to allow learning large-grid size possible and is able to perform training and inference on grid size of at least millions, two orders of magnitude higher than previous learning-based subsurface surrogate models. Experiments show that our HGNS outperforms strong baselines by a large margin, and our hybrid design and multi-step objective both contribute to the improved performance. Compared to standard subsurface solvers, we achieve a 50\% speedup with 1.1 million cells. Our model is deployed in industry pipeline for speeding up simulations for well placement and production forecasting.

Future work includes extending our HGNS to more complicated Dual Porosity Dual Permeability (DPDP) subsurface models, where fractures act as conduits for fast fluid flow. Based on our learned simulator, accelerated history matching can also be performed, where static properties can be inferred and updated by solving the inverse problem, conditioned to observed dynamic data.

\bibliographystyle{ACM-Reference-Format}
\vspace{-2.05mm}
\bibliography{reference}

\newpage
  
\newpage
\appendix

\onecolumn
\section{Appendices}
\subsection{Features Encoded}
\label{app:features}
Table \ref{tab:feature_table} shows all features we used for the experiments, consisting of dynamic, static, computed and control features.
\begin{table}[h!]
\centering
\caption{Encoded dynamic, static, computed and control features for our HGNS model and compared models. Here the node type one-hot encoding denotes whether a cell is a normal cell, injector, producer. The boundary encoding is a 3-vector encoding if a cell is near the boundary of the full grid, and has value ramping from 0 to 1 if it is from 5 to 1 cell distance from the boundary.}

\begin{tabular}{l|l|l|l}
\Xhline{2\arrayrulewidth}
\makecell{Dynamic features $X^t$} & \makecell{Static features $Q$} & \makecell{Computed features $R(X^t)$} & \makecell{Control features $U^t$} \\
\hline
Pressure P & Depth of cell &  Water relative permeability $k_{rw}(S_w)$ & Water injection rate $q_{w,inj}^t$ \\
Water volume $V_w$ & Porosity $\phi$ &  Oil relative permeability $k_{ro}(S_o)$ & Pressure at the well injector location\\
Oil volume $V_o$ & Pore volume $V$ & Spatial gradient of dynamic features $\nabla X^t$ & \\
& Connate water volume $V_{wc}$ &  & \\
& Permeability in $x$ direction $k_x$ &  & \\
& Permeability in $y$ direction $k_y$ &  & \\
& Permeability in $z$ direction $k_z$ &  & \\
& Transmissibility in $x$ direction $T_x$ &  & \\
& Transmissibility in $y$ direction $T_y$ &  & \\
& Transmissibility in $z$ direction $T_z$&  & \\
& Node type one-hot encoding &  & \\
& Boundary encoding &  & \\
\Xhline{2\arrayrulewidth}
\end{tabular}
\label{tab:feature_table}
\end{table}

\subsection{Hyperparameters for HGNS}
\label{app:hyperparameter}

Table \ref{tab:hyper_hgns} shows the hyperparameter values used for HGNS.
\begin{table*}[h!]
\centering
\caption{Hyperparameters used for HGNS}
\begin{tabular}{@{}l|l@{}}
\Xhline{2\arrayrulewidth}

Parameter name & Value \\
\hline
Number of GNLayers for the processor & 2 \\
Latent size for the processor & 16 \\    
Activation & elu \\
Type of nomalization & Group normalization \\
Number of layers for each MLP & 2 \\
Convoluion type & GNLayer \\
Number of neurons for each layer of MLP in the processor & 128 \\
Number of neurons for encoder MLP & 128 \\
Number of neurons for decoder MLP & 128 \\
Number of layers for encoder MLP & 2 \\
Number of layers for decoder MLP & 2 \\
Number of layers for the pooling and unpooling models & 1 \\
Number of groups for GroupNorm & 2 \\
Residual connection to use in GNN model & None \\
Fluid decoder model & MLP \\
Number of feature maps for first conv layer of U-Net encoder & 32 \\
Number of levels in the U-Net encoder/decoder path & 3 \\
Number of groups in U-Net group norm & 2\\

\Xhline{2\arrayrulewidth}
\end{tabular}
\label{tab:hyper_hgns}
\end{table*}

Table \ref{tab:hgns_press_enc_structure} shows the detailed structure of the HGNS pressure model's encoder. The initial input $x$ is forward propagated through each of the encoders in a sequential order from Encoder(0) $\rightarrow$ Encoder(1) $\rightarrow$ Encoder(2). Table \ref{tab:hgns_press_decf_structure} shows the structure of the HGNS pressure model from the first decoder layer to the final convolution output layer. The output from Encoder(2) will be forward propagated in the order of Decoder(0) $\rightarrow$ Decoder(1) $\rightarrow$ Final convolution.

\begin{table*}[h!]
\centering
\caption{HGNS pressure model encoder structure}
\begin{tabular}{l l l}
\Xhline{2\arrayrulewidth}
Encoder(0) & Encoder(1) & Encoder(2) \\
\hline
Input: $x$ & Input: $x$ & Input: $x$ \\
 & $x=\text{MaxPool3d}(2,2,2)(x)$ & $x=\text{MaxPool3d}(2,2,2)(x)$ \\

$x = \text{Conv3d}((3 \times 3 \times 3), 33, 32)(x)$ & $x = \text{Conv3d}((3 \times 3 \times 3), 32, 32)(x)$ & $x = \text{Conv3d}((3 \times 3 \times 3), 64, 64)(x)$\\
$x=\text{ReLU}(x)$  & $x=\text{ReLU}(x)$ & $x=\text{ReLU}(x)$ \\
$x=\text{GroupNorm(2, 32)}(x)$ & $x=\text{GroupNorm(2, 32)}(x)$ & $x=\text{GroupNorm(2, 64)}(x)$ \\
$x = \text{Conv3d}((3 \times 3 \times 3), 32, 32)(x)$ & $x = \text{Conv3d}((3 \times 3 \times 3), 32, 64)(x)$ & $x = \text{Conv3d}((3 \times 3 \times 3), 64, 128)(x)$\\
$x=\text{ReLU}(x)$ & $x=\text{ReLU}(x)$ & $x=\text{ReLU}(x)$ \\
$x=\text{GroupNorm(2, 32)}(x)$ & $x=\text{GroupNorm(2, 64)}(x)$ & $x=\text{GroupNorm(2, 128)}(x)$ \\

\Xhline{2\arrayrulewidth}
\end{tabular}
\label{tab:hgns_press_enc_structure}
\end{table*}

\begin{table*}[h!]
\centering
\caption{HGNS pressure model decoder and final convolution structure}
\begin{tabular}{l l l}
\toprule
Decoder(0) & Decoder(1) & Final convolution \\
\hline
Input: $x$ & Input: $x$ & Input: $x$ \\
$x=\text{InterpolateUpsampling}(x)$ & $x=\text{InterpolateUpsampling}(x)$ & $x=\text{Conv3d}((1 \times 1 \times 1), 32, 1)$ \\
$x = \text{Conv3d}((3 \times 3 \times 3), 196, 64)(x)$ & $x = \text{Conv3d}((3 \times 3 \times 3), 96, 32)(x)$ & \\
$x=\text{GroupNorm(2, 64)}(x)$ & $x=\text{GroupNorm(2, 32)}(x)$ & \\
$x = \text{Conv3d}((3 \times 3 \times 3), 64, 64)(x)$ & $x = \text{Conv3d}((3 \times 3 \times 3), 32, 32)(x)$ & \\
$x=\text{ReLU}(x)$ & $x=\text{ReLU}(x)$ & \\
$x=\text{GroupNorm(2, 64)}(x)$ & $x=\text{GroupNorm(2, 32)}(x)$ & \\

\bottomrule
\end{tabular}
\label{tab:hgns_press_decf_structure}
\end{table*}

Table \ref{tab:hgns_fluid_structure} shows the general structure of the HGNS fluid model structure, as well as the detailed structure for the node level MLP ($\text{MLP}_v$) and edge level MLP ($\text{MLP}_e$). The structure of $\text{MLP}_v$ and $\text{MLP}_e$ in the two GNLayers are the same.

\begin{table*}[h!]
\centering
\caption{HGNS fluid model structure}
\begin{tabular}{l l l}
\toprule
General structure & Node MLP ($\text{MLP}_v$) & Edge MLP ($\text{MLP}_e$) \\
\hline
Input: $x$ & Input: $x$ & Input: $x$ \\
$x=\text{ELU}(x)$ & $x=\text{Linear}(144, 128)(x)$ & $x=\text{Linear}(36, 128)(x)$\\
$x=\text{GNLayer}_0(x)$ & $x=\text{ELU}(x)$ & $x=\text{ELU}(x)$\\
$x=\text{GroupNorm(2, 16)}(x)$ & $x=\text{Linear}(128, 16)(x)$ & $x=\text{Linear}(128, 128)(x)$ \\
$x=\text{GNLayer}_1(x)$ & & \\

\bottomrule
\end{tabular}
\label{tab:hgns_fluid_structure}
\end{table*}

\vspace{2mm}
\begin{multicols}{2}
\subsection{Details for training}
Table \ref{tab:hyper_train} shows the values of all hyperparameters used for training. The noise added refers to the random walk noise added during training. As explained by \cite{sanchez2020learning}, adding random walk noise brings the training distribution closer to the distribution during prediction rollout. The weight of cell refers to a weight we assign to different cells depending on their distance to a well location. In cases where we want more accurate prediction results near well locations, we can assign a higher weight to the cell near the well, and a gradually decreasing weight for cells further away following some density function (e.g. Gaussian function).

\subsection{Details for dataset and pre-processing}
The datasets we used for training consist of 16 different trajectories, and the test set consists of 4 different trajectories. Each trajectory is the evolution of a $85 \times 220 \times 60$ grid, in the depth (vertical), length and width direction, respectively, and spans over 61 times steps. Such grid size amounts to a total of 1122000 cells. 

We create an edge in the graph between edge valid cell and its valid neighbor. If either the cell or its neighbor is an invalid cell, no edge will be created between this cell pair.

There can be a varying number of wells (producer or injector) in each trajectory, all between 5 and 10 in our dataset.

\vspace{3mm}
\begin{center}
\centering
\captionof{table}{Hyperparameters used for training}
\vspace{-1mm}
\begin{tabular}{@{}l|l@{}}
\Xhline{2\arrayrulewidth}

Parameter name & value \\
\hline
Loss function & MSE \\
Number of epochs & 40 \\
Batch size & 1 \\
Learning rate & 0.001 \\
Weight decay & 0 \\
Optimizer & Adam \\
Optimizer scheduler & cos \\
Noise added & 3e-5 \\
Weight of cell & \makecell[l]{Gaussian decay with std.=10, \\ minimum weight=0.2 and \\ weight=50 at well centers }  \\

\Xhline{2\arrayrulewidth}
\end{tabular}
\label{tab:hyper_train}
\end{center}
\end{multicols}

\end{document}